\def\year{2021}\relax
\documentclass[letterpaper]{article} % DO NOT CHANGE THIS
\usepackage{aaai21}  % DO NOT CHANGE THIS
\usepackage{times}  % DO NOT CHANGE THIS
\usepackage{helvet} % DO NOT CHANGE THIS
\usepackage{courier}  % DO NOT CHANGE THIS
\usepackage[hyphens]{url}  % DO NOT CHANGE THIS
\usepackage{graphicx} % DO NOT CHANGE THIS
\usepackage{natbib}  % DO NOT CHANGE THIS AND DO NOT ADD ANY OPTIONS TO IT
\usepackage{caption} % DO NOT CHANGE THIS AND DO NOT ADD ANY OPTIONS TO IT
\usepackage{booktabs} % To thicken table lines

\usepackage[english]{babel}
\usepackage{moresize}
\usepackage{amsmath}

\usepackage{algorithmic} 
\usepackage{balance}
\usepackage{comment}
\usepackage{paralist}
\usepackage{multirow}

\usepackage{filecontents}

\usepackage[english]{babel}
\usepackage[latin1]{inputenc}
\usepackage{mathrsfs}
\usepackage{graphicx}
\usepackage{amssymb}
\usepackage{url}
\usepackage{subfigure}
\usepackage{amsmath}
\usepackage{enumitem}
\usepackage[linesnumbered,algoruled,boxed,lined]{algorithm2e}
\usepackage{adjustbox}
\usepackage{amssymb}
\usepackage{times}

\usepackage{pgfplots}
\usetikzlibrary{pgfplots.dateplot}

\usepackage{filecontents}
\definecolor{tblue}{RGB}{31,119,180}
\definecolor{torange}{RGB}{255,127,14}
\definecolor{tgreen}{RGB}{44,160,44}
\definecolor{tred}{RGB}{214,39,40}
\definecolor{tpurple}{RGB}{148,103,189}

\usepackage{filecontents}

\newcommand{\hide}[1]{} %hide
\newcommand{\etal}{\textit{et al}.}

\newcommand{\ie}{\textit{i}.\textit{e}.}
\newcommand{\eg}{\textit{e}.\textit{g}.} 
\newcommand{\wrt}{\textit{w}.\textit{r}.\textit{t}} 
\newtheorem{Dfn}{Definition}

\title{Traffic Flow Forecasting with Spatial-Temporal Graph Diffusion Network}

\author{
Xiyue Zhang$^1$, Chao Huang$^2$\thanks{Corresponding author: Chao Huang}, Yong Xu$^{1,3,4}$, Lianghao Xia$^1$, Peng Dai$^2$ \\\Large{\bf Liefeng Bo$^2$, Junbo Zhang$^{5,6}$, Yu Zheng$^{5,6}$}\\
}
\affiliations{
    \textsuperscript{\rm 1}South China University of Technology, China\\
    \textsuperscript{\rm 2}JD Finance America Corporation, USA \\
    \textsuperscript{\rm 3}Communication and Computer Network Laboratory of Guangdong, China\\
    \textsuperscript{\rm 4}Peng Cheng Laboratory, China\\
    \textsuperscript{\rm 5}JD Intelligent Cities Research, China, \textsuperscript{\rm 6}JD Intelligent Cities Business Unit, JD Digits, China\\
    \{zhang.xiyue,cslianghao.xia\}@mail.scut.edu.cn, yxu@scut.edu.cn, chaohuang75@gmail.com\\
\{peng.dai,liefeng.bo\}@jd.com, \{msjunbozhang,msyuzheng\}@outlook.com
}

\def\model{ST-GDN}

\begin{document}

% \author{Anonymous Author(s)}

\maketitle

\begin{abstract}
Accurate forecasting of citywide traffic flow has been playing critical role in a variety of spatial-temporal mining applications, such as intelligent traffic control and public risk assessment. While previous work has made significant efforts to learn traffic temporal dynamics and spatial dependencies, two key limitations exist in current models. First, only the neighboring spatial correlations among adjacent regions are considered in most existing methods, and the global inter-region dependency is ignored. Additionally, these methods fail to encode the complex traffic transition regularities exhibited with time-dependent and multi-resolution in nature. To tackle these challenges, we develop a new traffic prediction framework--\underline{\textbf{S}}patial-\underline{\textbf{T}}emporal \underline{\textbf{G}}raph \underline{\textbf{D}}iffusion \underline{\textbf{N}}etwork (ST-GDN). In particular, \model\ is a hierarchically structured graph neural architecture which learns not only the local region-wise geographical dependencies, but also the spatial semantics from a global perspective. Furthermore, a multi-scale attention network is developed to empower \model\ with the capability of capturing multi-level temporal dynamics. Experiments on several real-life traffic datasets demonstrate that \model\ outperforms different types of state-of-the-art baselines. Source codes of implementations are available at \emph{https://github.com/jill001/ST-GDN}. 
\end{abstract}

% After that, we employ GRU to encode the temporal information in the time series data. 

\section{Introduction}
\label{sec:intro}

Accurate forecasting of traffic flow across different geographical regions in a city, have played a critical role in smart transformation systems, such as intelligent transportation~\cite{wei2018intellilight,huang2020cross} and public risk assessment~\cite{gao2019incomplete,2018deep}. For example, in disaster control, by predicting future traffic volume, local governments and communities is able to design better transportation scheduling and mobility management strategies, to mitigate the tragedies caused by the crowd flow~\cite{zhao2017feature}. In general, the objective of traffic prediction is to forecast the traffic volume (\eg, inflow and outflow of each region), from past traffic observations~\cite{diao2019dynamic}.

With the advancement of deep learning techniques, many efforts have been devoted to developing traffic prediction methods with various neural network architecture for spatial-temporal pattern modeling. Inspired by the sequence learning paradigm, recent neural networks have been utilized to model temporal effects of traffic variations~\cite{liu2016predicting,yu2017deep}. To make use of spatial features, some research work propose to adopt convolutional neural network to model correlations between adjacent regions~\cite{zhang2017deep}, along with using recurrent neural layers on the temporal dimension~\cite{yao2018deep}. Although both spatial and temporal correlations have been considered in existing methods, several key challenges have not been well addressed.

In real-life scenarios, traffic flow pattern is often complex and multi-periodic~\cite{zhang2017deep,deng2016latent}, as different views with respect to time resolutions (\eg, hourly, daily, weekly) reflect the traffic dynamics from different temporal dimensions. The captured temporal patterns are often complementary with each other~\cite{wu2018restful}. Hence, learning accurate representations of traffic variation patterns requires the collaboration of multiple views with different time resolutions. While recurrent neural network-based approaches have achieved good performance on various spatial-temporal sequence prediction tasks, they can only be effective for short-term, smooth dynamics and can hardly make predictions over the high-order multi-dimensional time horizons. 

Most current forecasting approaches merely focus on modeling nearby geographical correlations~\cite{yao2018deep,zhang2017deep}, while ignoring the cross-region inter-dependencies under a global context. For example, two geographical areas with similar urban functions (\eg, shopping zone or transportation hub) can be correlated in terms of their traffic distribution, although they are not spatially adjacent or even far away from each other~\cite{shen2018stepdeep,wang2017region}. Hence, the learned region-wise relational structures without the global-level traffic transition information, are insufficient to distill not only local geographical dependencies, but also global relations across regions, which leads to suboptimal predictions results.

To tackle the above challenges, we propose a new predictive framework \underline{\textbf{S}}patial-\underline{\textbf{T}}emporal \underline{\textbf{G}}raph \underline{\textbf{D}}iffusion \underline{\textbf{N}}etwork (\model), for region-specific traffic flow. In \model, we develop a multi-scale self-attention network to investigate multi-grained temporal dynamics across various time resolutions, in order to encode temporal hierarchy of traffic transitional regularities. To promote the collaboration of different granularity-aware temporal representations, an aggregation layer is proposed to model the underlying dependencies across multi-level temporal dynamics. In addition, the developed hierarchical graph neural network via attentive graph diffusion paradigm, endows the \model\ with the capability of incorporating spatial semantics from local-level spatially adjacent relations to global-level traffic pattern representations across the city in a joint way.

We highlight the key contributions of this work as below:
\begin{itemize}[leftmargin=*]
\item We highlight the critical importance of explicitly exploring the multi-resolution traffic transitional information and local-global cross-region dependencies, in studying the traffic prediction problem.

\item We propose a new traffic prediction framework (\model) which explicitly embeds multi-level temporal contextual signals into granularity-aware latent representations, with the cooperation of the designed multi-scale self-attention network and temporal hierarchy aggregation layer.

\item \model\ preserves both local and global region-wise dependencies, via a hierarchically structured graph neural architecture which is consisted of a graph attention network and convolution-based graph diffusion mechanism.

\item Our extensive experiments on three real-world datasets demonstrate that \model\ outperforms baselines of different types in yielding better forecasting performance. Furthermore, model efficiency study is conducted for \model\ in the traffic prediction process.
\end{itemize}

\section{Problem Definition}
\label{sec:model}

In this section, we begin with some key definitions and preliminary terms which are relevant to the solution.
%Then, we present our studied task of traffic prediction.

\begin{Dfn}
\textbf{Spatial Region}. We partition a city into $I\times J$ disjoint grids (given the geographical coordinates), in which each grid is regarded as a spatial region $r_{i,j}$ ($i\in [1,...,I]$, $j\in [1,...,J]$). $r_{i,j}$ is our target unit for traffic prediction.
\end{Dfn}

\begin{Dfn}
\textbf{Traffic Flow Tensor}. After the grid-based partition, we represent the citywide traffic volume distributions across regions during past $T$ time slots as a three-way tensor: $\textbf{X}\in \mathbb{R}^{I\times J\times T}$, where each entry $x_{i,j}^t$ denotes the traffic volume measurement at region $r_{i,j}$ in the $t$-th time slot (\eg, hour or day). To study the prediction on both the incoming and outgoing traffic follow, we generate two traffic flow tensors: $\textbf{X}^{\alpha}$ (incoming) and $\textbf{X}^{\beta}$ (outgoing), respectively.   
\end{Dfn}

\noindent \textbf{Task Formulation}. Based on the aforementioned definitions, the traffic prediction problem is formulated as: \textbf{Input}: the observed traffic volume information during past $T$ time slots across the entire city $\textbf{X}^{\alpha} \in \mathbb{R}^{I\times J\times T}$ and $\textbf{X}^{\beta} \in \mathbb{R}^{I\times J\times T}$. \textbf{Output}: a predictive function which effectively infers the unknown future traffic volume of regions.

\section{Methodology}
\label{sec:solution}

\begin{figure*}
    \centering
    %\vspace{-0.1in}
    \includegraphics[width=0.95\textwidth]{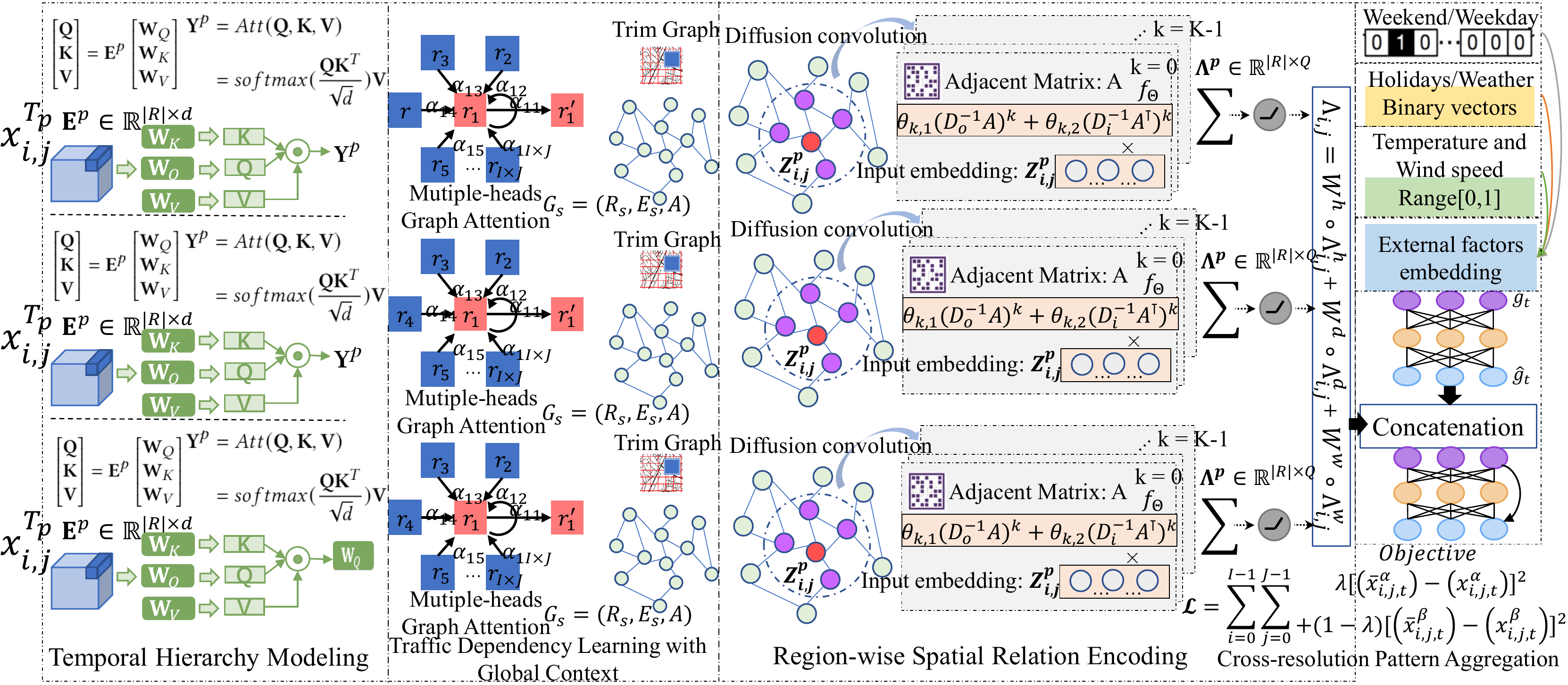}
    \vspace{-0.1in}
    \caption{The framework of our developed spatial-temporal graph diffusion networks.}
    \label{fig:fra}
    \vspace{-0.15in}
\end{figure*}

This section presents the our \model\ with the descriptions of different components (as shown in Figure~\ref{fig:fra}).\vspace{-0.1in}

\subsection{Temporal Hierarchy Modeling}
We first propose a multi-scale self-attention network to jointly map multi-level temporal signals into common latent representations, for capturing the complex traffic patterns.

\begin{Dfn}
\textbf{Temporal Resolution $p$}. We define $p$ to indicate how often we sample traffic volume measurement $x_{i,j}^t$ from the overall traffic flow tensor $\textbf{X}$, \ie, the time difference between two consecutive data points $x_{i,j}^t$ and $x_{i,j}^{t'}$ measured from region $r_{m,n}$. For example, $(t'-t)$ can be a hour, a day or a week, given the resolution $p$ is set as hourly, daily and weekly, respectively, \ie, $p\in \{hour,day,week\}$.
\end{Dfn}

Given each temporal resolution $p$, we could generate resolution-aware traffic series $\textbf{x}_{i,j}^{T_p}$, where $T_p$ is the corresponding traffic series length with the resolution of $p$. Then, we propose a self-attentive network to encode the traffic variation patterns from the temporal dimension. In particular, our encoder is built upon the scaled dot-product attention architecture with three transformation matrices: query ($\textbf{Q}\in \mathbb{R}^{T_p\times d}$), key ($\textbf{K}\in \mathbb{R}^{T_p\times d}$) and value ($\textbf{V}\in \mathbb{R}^{T_p\times d}$) matrices. The resolution-aware attentive aggregation mechanism can be formally presented with the matrix calculation:
\begin{align}
\textbf{Q} = \textbf{E}^p \cdot \textbf{W}_Q, \textbf{K} = \textbf{E}^p \cdot \textbf{W}_K, \textbf{V} = \textbf{E}^p \cdot \textbf{W}_V.
\end{align}
% \begin{small}
% \begin{align}
% \begin{bmatrix}
% \textbf{Q} \\ \textbf{K} \\ \textbf{V} 
% \end{bmatrix}
% = \textbf{E}^p 
% \begin{bmatrix}
% \textbf{W}_Q \\ \textbf{W}_K \\ \textbf{W}_V 
% \end{bmatrix}
% ;\textbf{Y}^p =\sigma(\frac{\textbf{Q}\textbf{K}^T}{\sqrt{d}})\textbf{V}
% \end{align}
% \end{small}
\noindent We further perform the operation of $\textbf{Y}^p =\sigma(\frac{\textbf{Q}\textbf{K}^T}{\sqrt{d}})\textbf{V}$, where $\textbf{y}_{i,j}^p \in \textbf{Y}^p$ and $\textbf{y}_{i,j}^p \in \mathbb{R}^{d}$ denotes the learned resolution-aware hidden representation of region $r_{i,j}$. $E_p \in \mathbb{R}^{|R|\times d}$ is the initialized embeddings of all regions $r_{i,j}\in R$. Additionally, $\sigma(\cdot)$ denotes the softmax function. Here, $\textbf{W}_Q$, $\textbf{W}_K$, $\textbf{W}_V$ are projection matrices.

\subsection{\bf Traffic Dependency Learning with Global Context}
The goal of this step is to exploit the global-level dependencies across different regions in terms of their dynamic traffic transition patterns. Towards this end, we first define a region graph $G=(R,E)$, in which $R$ is the region set and $E$ denotes the pairwise relationships between two spatial regions. Motivated by the attention-based neural network in encoding the dependencies among regions~\cite{2019mist}, we develop an attentive aggregation mechanism to capture both local and global traffic dependency between regions. Specifically, we perform the message aggregation over $G$ with the following attentive operations.
\begin{align}
m_{(i,j)\leftarrow (i',j')}^p = \mathop{\Bigm|\Bigm|}\limits_{h=1}^{H} \omega_{(i,j);(i',j')}^h \cdot \textbf{Y}^p \cdot \textbf{W}^p
\end{align}
\noindent where $m_{(i,j)\leftarrow (i',j')}^p$ is the feature message propagated from region $r_{i',j'}$ to $r_{i,j}$. Here, we endow the cross-region relevance encoding with multi-head ($h\in [1,...,H]$), to capture the region-wise relation semantic from different learning subspaces. Furthermore, $\textbf{W}^p \in \mathbb{R}^{d\times d}$ is the parameterized projection matrix. The underlying attentive relevance $\omega_{(i,j);(i',j')}^h$ is formally estimated as follows:
\begin{small}
\begin{align}
\omega_{(i,j);(i',j')}^h = \dfrac{exp(LR(\boldsymbol{\alpha}^T [\widetilde{\textbf{y}}^p_{i,j} || \widetilde{\textbf{y}}^p_{i',j'}]))}{\sum_{(i',j')\in \mathcal{N}(i,j)} exp(LR(\boldsymbol{\alpha}^T [ \widetilde{\textbf{y}}^p_{i,j} || \widetilde{\textbf{y}}^p_{i',j'}]))} \nonumber
\label{equ:attention coefficient}
\end{align}
\end{small}
\noindent where we perform concatenation between $\widetilde{\textbf{y}}^p_{i',j'}$ and $\widetilde{\textbf{y}}^p_{i',j'}$ ($\widetilde{\textbf{y}}^p_{i',j'} = {\textbf{y}}^p_{i',j'} \cdot \textbf{W}^p$). Then, the attentive coefficient vector $\alpha$ is incorporated with the production. $LR(\cdot)$ denotes the \emph{LeakyReLU} function. Based on the constructed message and learned quantitative region-wise relevance score $\omega_{(i,j);(i',j')}$, we perform the information aggregation as:
\begin{align}
\textbf{z}_{i,j}^p=f(\sum_{r_{i',j'}\in\mathcal{N}_{i,j}} m_{(i,j)\leftarrow (i',j')}^p)
\end{align}
\noindent where $\textbf{z}_{i,j}^p$ is the aggregated feature embedding of $r_{i,j}$.

\noindent \textbf{High-order Information Propagation}. The information aggregation from the $(l)$-th layer to the $(l+1)$-th layer with the high-order relation modeling is represented as:
\begin{align}
\textbf{z}_{i,j}^{p,(l+1)} \leftarrow \mathop{\textbf{Aggregate}}\limits_{i\in N_u(j); j'\in N_v(j)} \Big ( \textbf{Propagate}( \textbf{z}_{i,j}^{p,(l)}, G) \Big )
\end{align}
\noindent Propagate($\cdot$) and Aggregate($\cdot$) denotes the message construction and information fusion, respectively. We finally generate the global-level representation of region $r_{i,j}$ as: $\textbf{z}_{i,j}^{p} = \textbf{z}_{i,j}^{p,(l)} \oplus ... \oplus \textbf{z}_{i,j}^{p,(L)}$. $\oplus$ is the element-wise addition.

\subsection{\bf Region-wise Relation Learning with Graph Diffusion Paradigm}
In addition to the global dependencies across different regions in terms of their traffic evolving patterns, we further incorporate spatial relationships between regions into our prediction framework. Particularly, motivated by~\cite{li2017diffusion}, we develop a graph-structured diffusion network to refine the learned resolution-aware region representations $\textbf{z}_{i,j}^{p}$ from the above graph attention module. We generate another region-wise relation graph $G_s=(R_s, E_s, A)$ which jointly preserves the geographical adjacent relations ($r_{i,j}$'s $\sqrt{K}\times \sqrt{K}= K$ neighboring regions) and high traffic dependencies (larger $\omega_{(i,j);(i',j')}$ value). $A$ denotes the adjacent matrix weighted by a vertex distance function. Here, we define $D_o = \textbf{A} \cdot \textbf{I}$ to denote the out-degree diagonal matrix, where $\textbf{I}$ is the identify matrix of $G_s$. The designed diffusion convolution operation performs the diffusion process across nodes in $G_s$ to generate new feature representations:
\begin{align}
f( \textbf{z}_{i,j}^{p} )_{\Theta} = \sum_{k=0}^{K-1}(\theta_{k,1}(D^{-1}_{o} \textbf{A})^k+\theta_{k,2}(D^{-1}_{i} \textbf{A}^{\intercal})^k)\textbf{z}^{p}_{i,j}
\end{align}
\noindent where $\theta_{k,1}$, $\theta_{k,2}\in \mathbb{R}^{K\times 2}$. $D^{-1}_{o} \textbf{A}$ (in-degree) and $D^{-1}_{i} \textbf{A}^{\intercal})$ (out-degree) denote the bi-directional transition matrices of the diffusion process, which corresponds to the inflow and outflow in our prediction scenario. The parameter tensor denoted as $\Theta\in \mathbb{R}^{Q\times d\times K \times 2}$, in which the Q-dimensional output $\boldsymbol{\Lambda}^{p}\in \mathbb{R}^{|R|\times Q}$ of diffusion convolutional layer is given:
\begin{small}
\begin{align}
\boldsymbol{\Lambda}^{p}_{q}= LeakyReLU\big(\sum_{d'=1}^{d}f(\textbf{Z}^{p}_{d'})_{\Theta_{q, d'}} \big)
\end{align}
\end{small}
\noindent where $q\in\{1,..., Q\}$. The obtained region representation $\boldsymbol{\Lambda}^{p}_{i,j}$ jointly preserves the temporal (traffic time-varying patterns) and spatial (geographical relations) contextual signals under a global perspective.

% \subsection{Aggregation Layer over Temporal Hierarchy}
We next aggregate the resolution-aware traffic representation $\boldsymbol{\Lambda}^{p}_{i,j}$ by introducing a gating mechanism. To be specific, our gated aggregation mechanism conducts the parametric matrix-based sum operation over the multi-resolution traffic pattern representations, \ie, hourly ($\boldsymbol{\Lambda}^{p_h}$), daily ($\boldsymbol{\Lambda}^{p_d}$) and weekly ($\boldsymbol{\Lambda}^{p_w}$) as: $\boldsymbol{\Lambda}_{i,j}= \textbf{W}^{h} \circ \boldsymbol{\Lambda}^{h}_{i,j} + \textbf{W}^{d} \circ \boldsymbol{\Lambda}^{d}_{i,j} + \textbf{W}^{w} \circ \boldsymbol{\Lambda}^{w}_{i,j}$.
% \begin{align}
% \boldsymbol{\Lambda}_{i,j}= \textbf{W}^{h} \circ \boldsymbol{\Lambda}^{h}_{i,j} + \textbf{W}^{d} \circ \boldsymbol{\Lambda}^{d}_{i,j} + \textbf{W}^{w} \circ \boldsymbol{\Lambda}^{w}_{i,j}
% \label{equ:pooling}
% \end{align}
\noindent Here, the trainable transformation matrices are denoted as $\textbf{W}^{h}$, $\textbf{W}^{d}$ and $\textbf{W}^{w}$ corresponding to hourly, daily and weekly patterns. We finally generate the conclusive multi-resolution traffic representation $\boldsymbol{\Lambda}_{i,j}$ which preserves multi-grained temporal hierarchy of traffic regularities.

\subsection{\bf Traffic Prediction Phase}
In urban sensing, there exist external factors (\eg, meteorological conditions) which impact traffic transitional regularities. Thus, we further augment our \model\ with the capability of fusing external factors. In particular, we consider several types of external factors: Weather conditions, Temperature/$^\circ$C, Wind speed/mph. We map these features into vectors $\textbf{g}_t$. After that, we utilize a multi-layer perceptron framework to perform projection over $\hat{\textbf{g}}_t$. Finally, we feed the concatenated embedding ($\boldsymbol{\Lambda}_{i,j}$ and $\hat{\textbf{g}}_t$) into the prediction layer to infer the traffic volume of each region.\\

\noindent \textbf{Optimized Loss Function}. We define our loss function with the joint consideration of inflow and outflow traffic volume of each region in a city as below:
\begin{align}
\begin{split}
\mathcal{L}= \sum_{i=0}^{I-1}\sum_{j=0}^{J-1} \lambda[(\bar{x}^{\alpha}_{i,j,t})-(x^{\alpha}_{i,j,t})]^{2}\\ +(1-\lambda)[(\bar{x}^{
\beta}_{i,j,t})-(x^{\beta}_{i,j,t})]^{2}
\label{equ:loss}
\end{split}
\end{align}
\noindent where $\bar{x}^{\alpha}_{i,j,t}$ and $\bar{x}^{\beta}_{i,j,t}$ denote the estimated incoming and outgoing traffic volume of region $r_{i,j}$ at the $t$-th time slot, respectively. Their influences are decided by $\lambda$. Ground truth information are represented $x^{\alpha}_{i,j,t}$ and $x^{\beta}_{i,j,t}$.\\

\noindent \textbf{Model Complexity Analysis}. 
In this part, we analyze the time complexity of our \model\ framework. Particularly, the multi-scale self-attentive network takes $O(3\times T\times I\times J\times d)$ for learning query, key and value matrices, and $O(3\times T^2\times d)$ for weighted summation. The next attentional graph module takes $O(3\times I^2\times J^2\times d')$ to estimate the relevance scores and perform feature aggregation, which dominates the computational cost of our \model. Additionally, the graph diffusion-based spatial relation modeling component takes $O(K\times |E_s|)$ complexity.
%The external factor fusion takes the $O(3\times d)$ complexity.
\section{Evaluation}
\label{sec:eval}

\begin{table*}[t]
\centering
    %\scriptsize
    \normalsize
	\label{tab:Data statistic}
	\begin{tabular}{cccc|ccc}
		\hline
		Dataset           & BJ-Taxi    & NYC-Taxi & NYC-Bike & External Factors & Beijing & New York City\\
		\midrule
		Data type         &Taxi GPS    & Taxi GPS   & Bike Rental & Weather & sunny, rainy \etal & sunny, rainy \etal\\
		Time interval     &30 minutes  & 30 minutes & one hour& Temperature/$^\circ$C & $[-24.6, 41.0]$&$[-10.3, 31.40]$\\
		Gird map size &32$\times$32&10$\times$20& 16$\times$8& Wind speed/mph & $[0, 48.60]$&$[0, 63.75]$\\
% 		Time span&\begin{tabular}{c}
% 			7/1/2013-10/30/2013\\
% 		    3/1/2014-6/30/2014\\
% 		    3/1/2015-6/30/2015\\
% 			11/1/2015-4/10/2016\\
% 		\end{tabular}&4/1/2014-9/30/2014&1/1/2015-3/1/2015&7/1/2016-8/29/2016\\
		\# of records&34,000+&22,000,000+&6,800+& holidays & weekends, holidays & weekends, holidays\\
		\hline
	\end{tabular}
	\vspace{-0.1in}
	\caption{Statistical information of experimented datasets.}
	\vspace{-0.1in}
	\label{tab:data}
\end{table*}

\begin{figure*}
    \centering
    \normalsize
    %\vspace{-0.1in}
    \subfigure[][NYC-Taxi]{
        \centering
        \includegraphics[width=0.15\textwidth]{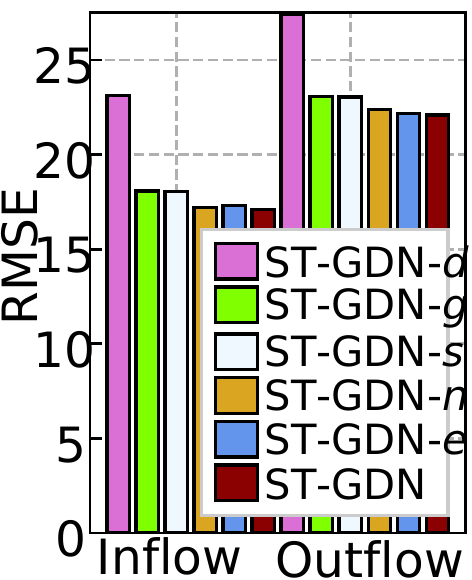}
        \label{fig:taxiNYC_RMSE}
        }
    \subfigure[][NYC-Taxi]{
        \centering
        \includegraphics[width=0.15\textwidth]{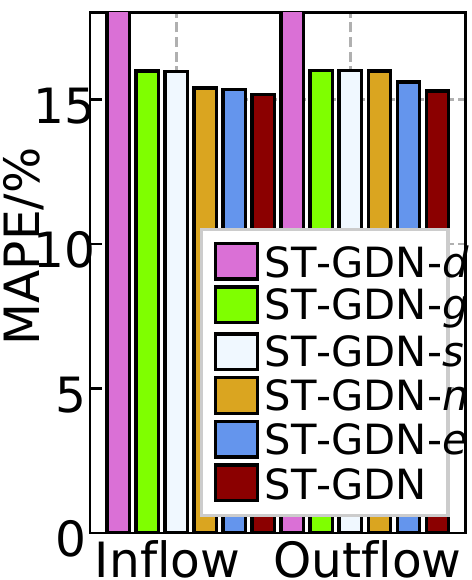}
        \label{fig:taxiNYC_MAPE}
        }
    \subfigure[][BJ-Taxi]{
        \centering
        \includegraphics[width=0.15\textwidth]{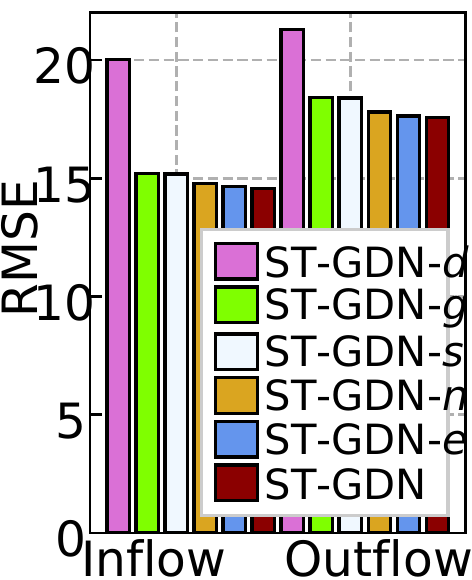}
        \label{fig:taxiBJ_RMSE}
        }
    \subfigure[][BJ-Taxi]{
        \centering
        \includegraphics[width=0.15\textwidth]{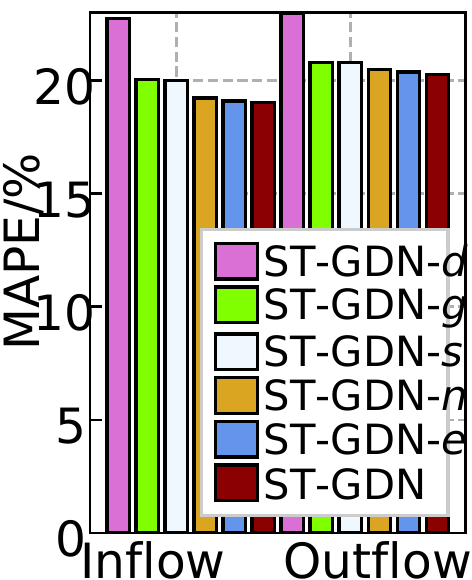}
        \label{fig:taxiBJ_MAPE}
        }
    \subfigure[][NYC-Bike]{
        \centering
        \includegraphics[width=0.15\textwidth]{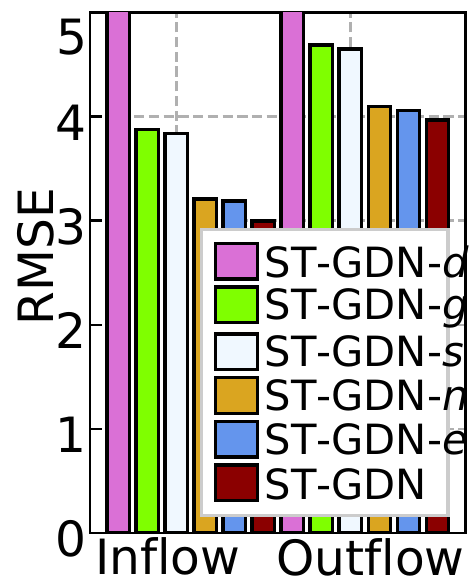}
        \label{fig:bikeNYC_RMSE}
        }
    \subfigure[][NYC-Bike]{
        \centering
        \includegraphics[width=0.15\textwidth]{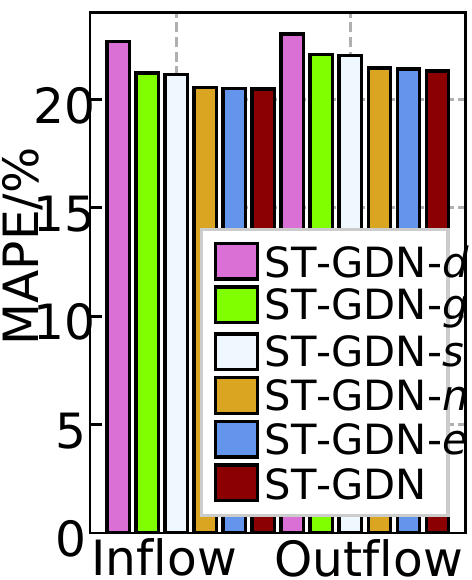}
        \label{fig:bikeNYC_MAPE}
        }
    \vspace{-0.1in}
    \caption{Model ablation study of ST-GDN framework in terms of RMSE and MAPE.}
    \label{fig:ablation}
    
\end{figure*}

In this section, we evaluate the performance of \emph{\model} on a series of experiments on several real-world datasets, which are summarized to answer the following research questions:
\begin{itemize}[leftmargin=*]
\item \textbf{RQ1}: How is the overall traffic prediction performance of \emph{\model} as compared to various baselines?
\item \textbf{RQ2}: How do designed different sub-modules contribute to the model performance?
\item \textbf{RQ3}: How does \emph{\model} perform \wrt\ different time granularity configurations for temporal context modeling?
% \item \textbf{RQ4}: What explainable relational patterns does \emph{\model} capture across geographical regions? 
\item \textbf{RQ4}: What is the influence of hyperparameter settings? 
\item \textbf{RQ5}: How is the model efficiency of \emph{\model}? 
\end{itemize}

\subsection{Experimental Settings}
\subsubsection{\bf Data Description.}
Experiments are performed on three real-world traffic datasets, which are summarized in Table~\ref{tab:data}:

\noindent \textbf{BJ-Taxi}~\cite{zhang2017deep}. There are 34,000+ processed taxi trajectories included in this data. Each trajectory is mapped into one of $32\times32$ grid-based geographical regions. The traffic volume is measured every half an hour.

\noindent \textbf{NYC-Taxi}~\cite{yao2019revisiting}. This data contains 22,000,000+ taxi trajectories collected from 01/01/2015 to 03/01/2015 in New York City with a $10\times 20$ grid map. The traffic data sample period is also half an hour.

\noindent \textbf{NYC-Bike}~\cite{zhang2017deep}. It includes the trajectories of the bike system from New York with a $16\times8$ grid map. Traffic volume is estimated on a hourly basis.\\\vspace{-0.12in}

\subsubsection{Evaluation Protocols.}
We leverage two representative metrics for evaluation: \emph{Root Mean Squared Error (RMSE)} and \emph{Mean Absolute Percentage Error (MAPE)}.
%~\cite{liang2019urbanfm}

\subsubsection{\bf Methods for Comparison.}
% In the performance comparison between our method and state-of-the-art traffic forecasting techniques, we consider the following baselines with various model structures.
In the comparison, we consider the following baselines with various model structures.

% \noindent \textbf{Traditional Time Series Prediction Approaches}:
\begin{itemize}[leftmargin=*]
\item \textbf{ARIMA}~\cite{icdm12}. it is a representative method for forecasting time series data.
\item \textbf{Support Vector Regression (SVR)}~\cite{chang2011libsvm}: another traditional time series analysis model via learning feature mapping functions.
% \end{itemize}

% \noindent \textbf{Conventional Hybrid Learning Approach}:
%\begin{itemize}[leftmargin=*]
\item \textbf{Fuzzy+NN}~\cite{srinivasan2009computational}: it integrates the feed-forward neural layers with the fuzzy input filter to model the traffic patterns. 
% \end{itemize}

% \noindent \textbf{Recurrent Spatial-Temporal Prediction Methods}:
%\begin{itemize}[leftmargin=*]
\item \textbf{RNN}~\cite{liu2016predicting}: it leverages the recurrent neural networks for capturing both the spatial and temporal effects for making sequential data prediction.
\item \textbf{LSTM}~\cite{yu2017deep}: it jointly models the normal and abnormal traffic variations based on stacked long short-term memory networks. 
% \end{itemize}

%\noindent \textbf{Convolution-based Network for Traffic Forecasting}:
% \begin{itemize}[leftmargin=*]
\item \textbf{DeepST}~\cite{zhang2016dnn}: it utilizes the convolution neural network to encode the spatial correlations between regions over a citywide grid map.
\item \textbf{ST-ResNet}~\cite{zhang2017deep}: the residual connection technique is employed to alleviate overfitting issue for spatial-temporal prediction. 
%\end{itemize}

%\noindent \textbf{Convolutional Recurrent Predictive Solution}:
%\begin{itemize}[leftmargin=*]
\item \textbf{DMVST-Net}~\cite{yao2018deep}: it integrates the graph embedding method with the joint convolutional recurrent networks to capture spatial-temporal signals
\item \textbf{DCRNN}~\cite{li2017diffusion}: it is a data-driven forecasting framework with diffusion recurrent
neural network to capture the spatial-temporal dependencies.
% \end{itemize}

%\noindent \textbf{Attentive Traffic Prediction Model}:
%\begin{itemize}[leftmargin=*]
\item \textbf{STDN}~\cite{yao2019revisiting}: it designs a periodically shifted attention for learning transition regularities of traffic. 
%\end{itemize}

%\noindent \textbf{Traffic Prediction with Graph Neural Networks}:
%\begin{itemize}[leftmargin=*]
\item \textbf{ST-GCN}~\cite{yubingspatio}: it is an integrative framework of graph convolution network and convolutional sequence modeling layer for modeling spatial and temporal dependencies.
\item \textbf{ST-MGCN}~\cite{geng2019spatiotemporal}: it develops a multi-modal graph convolutional network to capture region-wise non-Euclidean pair-wise correlations.
\item \textbf{GMAN}~\cite{zheng2020gman}: it is a encoder-decoder traffic prediction method based on the graph multi-attention.
%\end{itemize}

%\noindent \textbf{Deep Hybrid Traffic Flow Predictive Models}:
%\begin{itemize}[leftmargin=*]
\item \textbf{UrbanFM}~\cite{liang2019urbanfm}: it is a deep fusion network to model traffic flow distributions.
\item \textbf{ST-MetaNet}~\cite{pan2019urban}: it is a meta-learning approach to perform knowledge transfer across series with a recurrent graph attentive network.
\end{itemize}

\subsubsection{\bf Parameter Settings.}
The \emph{\model} is implemented with Tensorflow. The training phase is performed using the Adam optimizer with the learning rate of $1e^{-3}$ and batch size of 32. The embedding dimension size $d$ and the depth recursive graph neural layers $L$ are set as 64 and 3, respectively. We select the input sequence length from the range of $\{1, 2, 3, 4, 5, 6\}$, $\{1, 2, 3, 4, 5\}$, $\{1, 2, 3, 4, 5, 6\}$, which respectively corresponds to three different time resolutions (hour--$T_h$, day--$T_d$ and week--$T_w$). We stack three feed-forward layers in the final prediction phase. The experiments of most baselines are performed with their released code.

\subsection{Performance Comparison (RQ1)}
\begin{figure}
	\centering
	\subfigure[][ST-GCN]{
		\centering
		\includegraphics[width=0.14\textwidth]{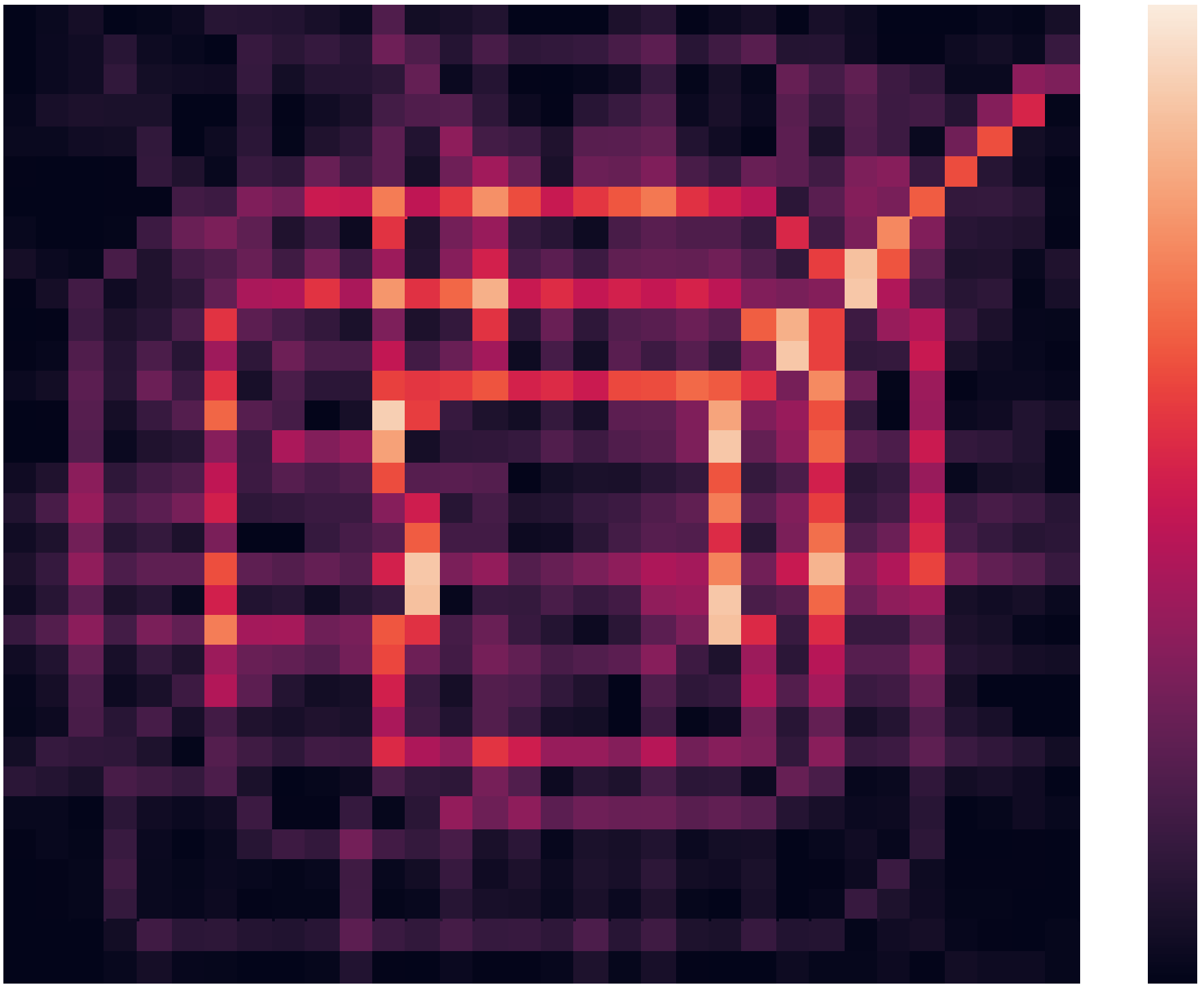}
		\label{fig:hdwtaxiNYC_MAPE}
	}
	\subfigure[][ST-MGCN]{
		\centering
		\includegraphics[width=0.14\textwidth]{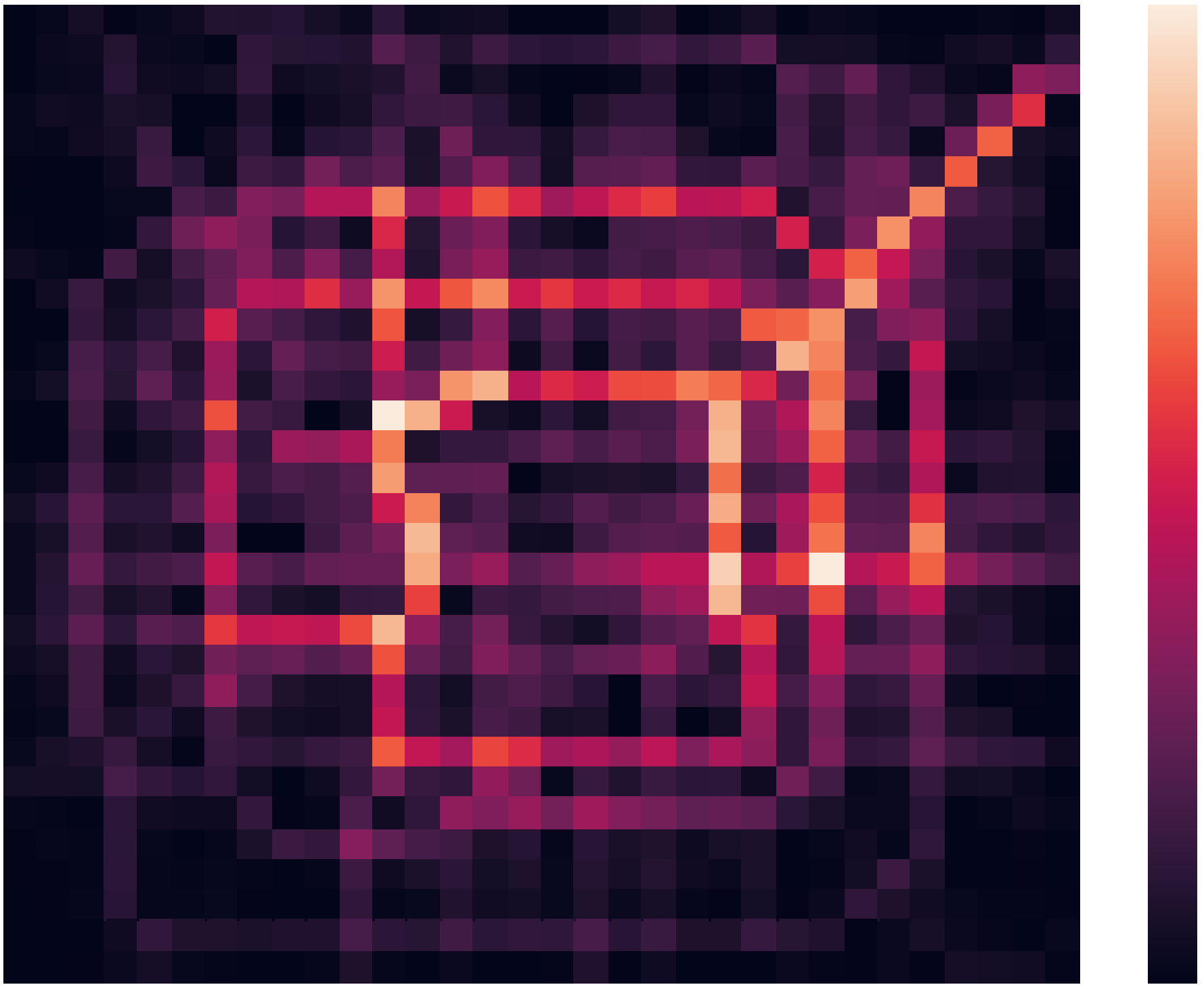}
		\label{fig:hdwtaxiBJ_RMSE}
	}
	\subfigure[][ST-MetaNet]{
		\centering
		\includegraphics[width=0.14\textwidth]{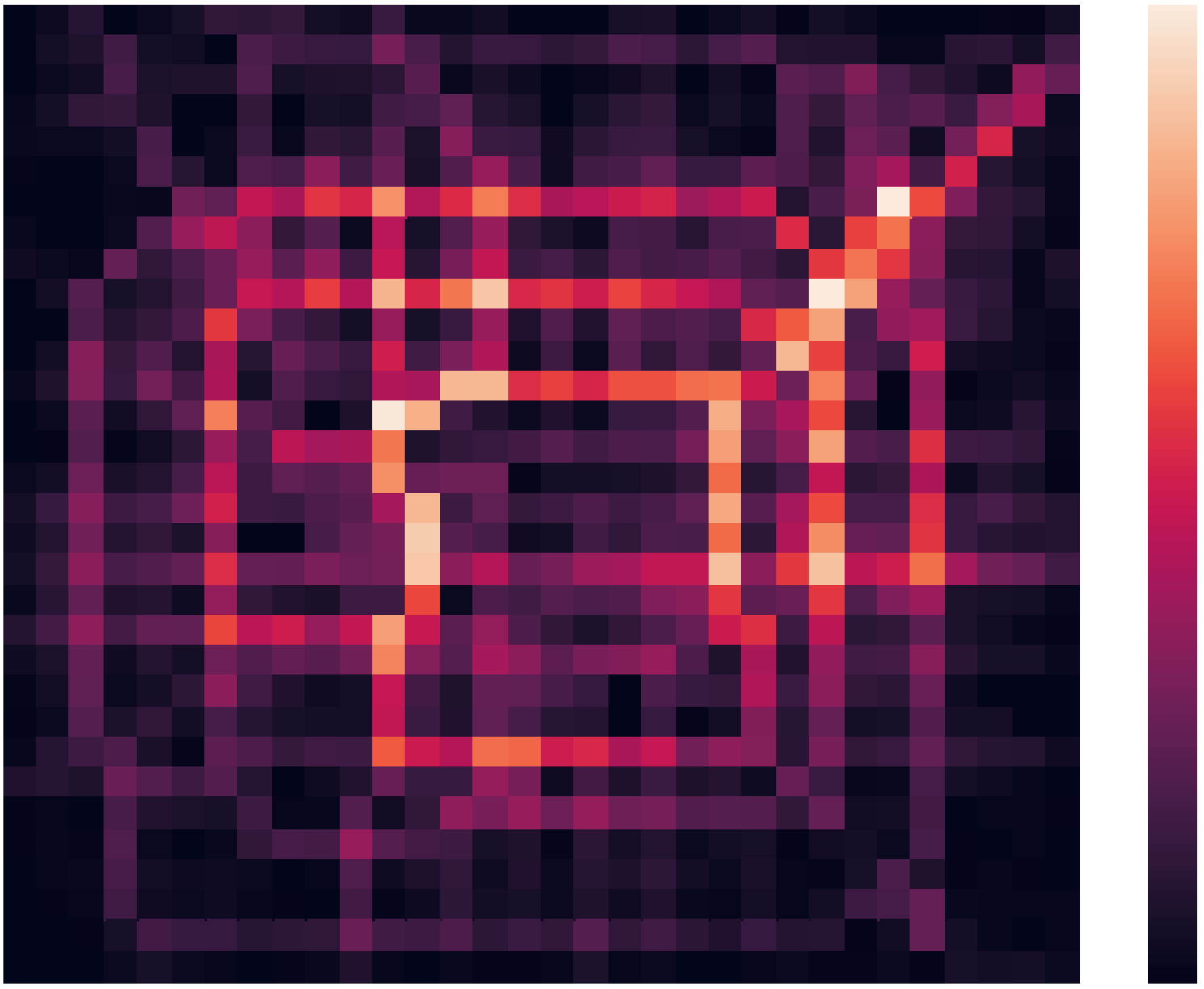}
		\label{fig:hdwtaxiBJ_MAPE}
	}
	\subfigure[][DCRNN]{
		\centering
		\includegraphics[width=0.14\textwidth]{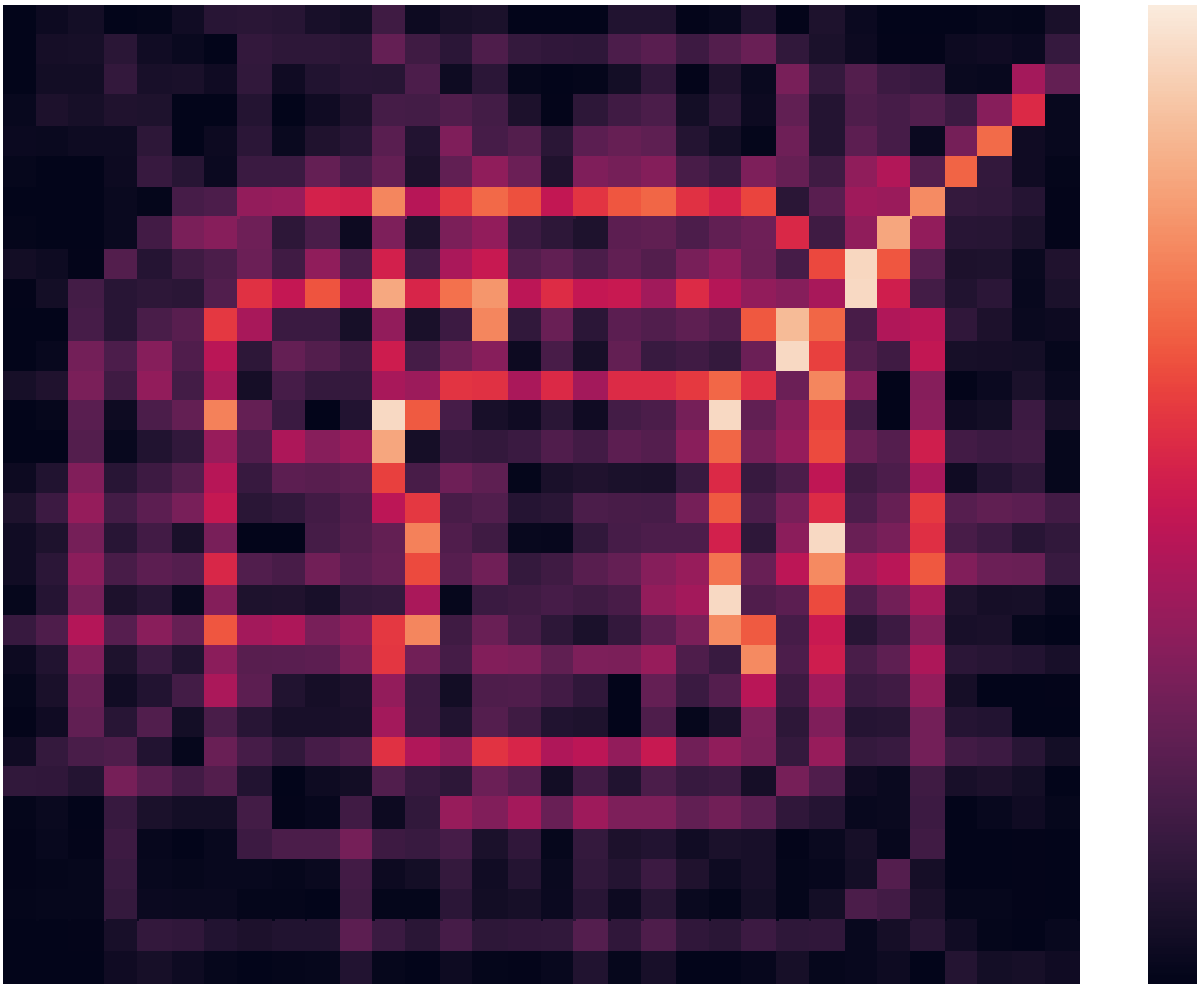}
		\label{fig:hdwbikeNYC_RMSE}
	}
	\subfigure[][GMAN]{
		\centering
		\includegraphics[width=0.14\textwidth]{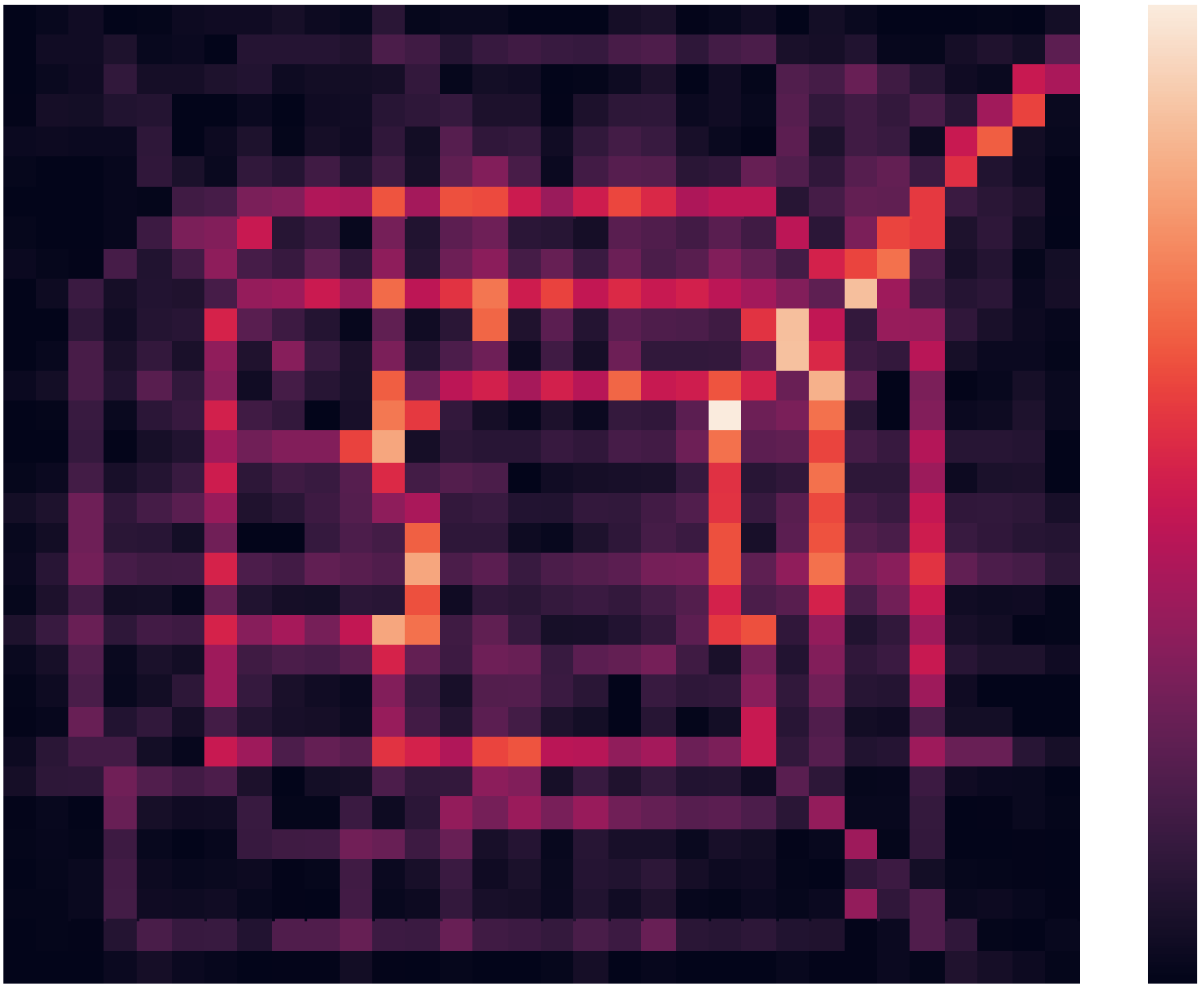}
		\label{fig:hdwbikeNYC_RMSE}
	}
	\subfigure[][ST-GDN]{
		\centering
		\includegraphics[width=0.14\textwidth]{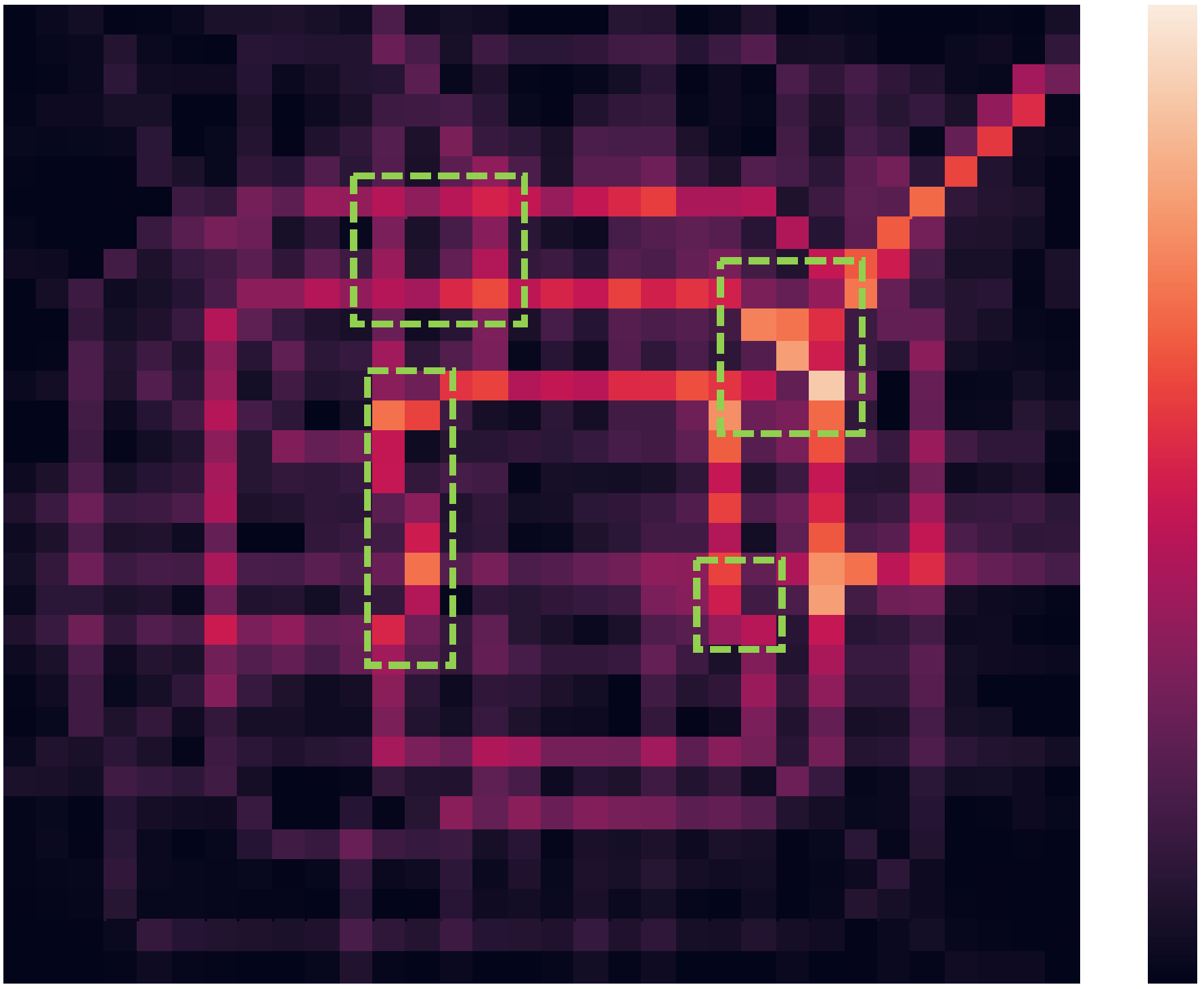}
		\label{fig:hdwbikeNYC_MAPE}
	}
	\caption{Visualization for Traffic Prediction Errors.}
	\label{fig:error}
\end{figure}

\noindent \textbf{Performance Superiority of \emph{\model}}. The comparison results of all methods are presented in Table~\ref{tab:results}. We can observe that \emph{\model} consistently yields the best performance in all cases, which demonstrates the effectiveness of our \emph{\model} in jointly modeling of multi-level temporal dynamics and global-level region-wise dependencies. Figure~\ref{fig:error} visualize the prediction error ($[(\bar{x}_{i,j,t})-(x_{i,j,t})]^{2}$) of our \emph{\model} and five best performed baselines on BJ-taxi data, where a brighter pixel means a larger error. The superiority of \emph{\model} can still be observed, which is consistent with the quantitative results in Table~\ref{tab:results}. \\

\noindent \textbf{Performance Comparison between Baselines}. Compared with conventional time series approaches, neural network-based models perform better in most evaluation cases. The subsequent attention-based and recurrent-convolutional network methods (\eg, STDN, DMVST-Net) obtain better performance than recurrent neural models (\eg, D-LSTM), which justifies the necessity to simultaneously capture both spatial and temporal relations in traffic prediction. Among various baselines, GNN-based methods have better performance than other types of competitors, which ascertains the rationality of designing graph-structured information aggregation mechanism to fuse spatial and temporal signals. 

\begin{table*}%[htbp]
    \centering
    %\footnotesize
    \normalsize 
	\label{tab:Exonold}
	\begin{tabular}{c|cc|cc|cc|cc|cc|cc}
		\hline
		Datasets&
		\multicolumn{4}{|c}{BJ-Taxi}&\multicolumn{4}{|c}{NYC-Bike}&\multicolumn{4}{|c}{NYC-Taxi}\\
		\cline{2-13} 
		Metrics&\multicolumn{2}{|c|}{RMSE}&\multicolumn{2}{c|}{MAPE (\%)}&\multicolumn{2}{c|}{RMSE}&\multicolumn{2}{c|}{MAPE (\%)}&\multicolumn{2}{c|}{RMSE}&\multicolumn{2}{c}{MAPE (\%)}\\
		\cline{2-13} 
		Methods&In&Out&In&Out&In&Out&In&Out&In&Out&In&Out\\
		\hline
		ARIMA&22.10&24.01&30.89&32.24&9.30&11.81&35.82&36.47&27.21&36.54&20.90&22.18\\
		SVR&21.44&22.12&22.64&22.32&8.65&9.07&23.58&24.10&26.16&34.71&18.25&21.01\\
		Fuzzy+NN&22.35&23.06&22.67&22.73&8.56&9.17&24.03&24.48&25.98&34.50&18.92&21.54\\
		RNN&27.16&27.90&24.17&24.72&8.99&9.24&28.22&28.58&29.88&37.23&25.97&26.55\\
		LSTM&26.99&27.56&23.64&24.17&8.64&9.10&27.51&28.07&29.52&37.04&25.81&26.11\\
		DeepST&19.30&21.06&22.45&22.52&7.66&8.16&22.81&23.21&23.56&26.79&22.34&22.39\\
		ST-ResNet&17.00&22.31&23.51&23.74&6.28&6.61&23.92&24.79&21.72&26.30&21.12&21.24\\
		DMVST-Net&16.61&17.14&22.52&23.06&5.82&6.09&22.45&23.67&20.63&25.80&17.19&17.44\\
		STDN&15.19&18.63&21.04&22.13&4.50&5.92&21.71&22.61&19.31&24.19&16.43&16.59\\
		UrbanFM&15.18&18.42&20.54&20.88&3.99&4.64&21.59&22.47&19.11&24.14&16.34&16.46\\
		ST-MetaNet&15.06&18.29&19.91&20.74&3.85&4.64&21.26&22.18&18.30&23.88&16.19&16.27\\
		DCRNN&15.13&18.37&20.14&20.88&3.86&4.65&21.14&21.05&18.19&23.74&16.11&16.16\\
		ST-GCN&15.11&18.30&19.92&20.77&3.76&4.70&21.12&21.94&18.02&23.08&15.94&15.92\\
		ST-MGCN&15.08&18.25&19.96&20.70&3.75&4.63&21.04&21.95&17.97&23.00&15.87&15.91\\
		GMAN&15.07&18.23&19.97&20.68&3.73&4.64&21.02&21.93&17.95&22.96&15.84&15.89\\
		\hline
		\textbf{\emph{\model}}&\textbf{14.57}&\textbf{17.56}&\textbf{19.03}&\textbf{20.27}&\textbf{3.00}&\textbf{3.97}&\textbf{20.48}&\textbf{21.31}&\textbf{17.10}&\textbf{22.09}&\textbf{15.17}&\textbf{15.29}\\
		\hline
	\end{tabular}
	\vspace{-0.1in}
	\caption{Performance comparison of all methods on three datasets in terms of \emph{RMSE} and \emph{MAPE}.}
	\vspace{-0.1in}
	\vspace{-0.1in}
	\label{tab:results}
\end{table*}

\subsection{Comparison with Variants (RQ2)}
We perform ablation experiments to analyze the effects of sub-modules in our \emph{\model} framework with five variants:
\begin{itemize}[leftmargin=*]
\item \emph{\model}-s: \emph{\model} without the multi-scale self-attention network to capture multi-level traffic dynamics.
\item \emph{\model}-g: \emph{\model} without the graph attention module to model the global region-wise traffic dependencies.
\item \emph{\model}-d: \emph{\model} without the graph diffusion network to integrate spatial context with cross-region traffic pattern correlations for representation recalibration.
\item \emph{\model}-n: \emph{\model} without the incorporation of neighborhood spatial context into the graph diffusion. 
\item \emph{\model}-e: \emph{\model} without the external factor fusion.
\end{itemize}

The evaluation results are shown in Figure~\ref{fig:ablation}. We can observe that the joint version of \emph{\model} outperforms other variants consistently. Hence, each designed sub-modules has positive effects for prediction performance improvement. It is necessary to build a joint framework to collectively integrate the multi-resolution traffic temporal patterns, global region-wise traffic dependencies, and regions' geographical relations, into the spatial-temporal traffic pattern modeling.

\subsection{Multi-Resolution Temporal Effects (RQ3)}
In this subsection, we study the effects of different temporal resolution settings in our integrative architecture of multi-scale self-attention network and cross-resolution pattern aggregation layer, with the following contrast models:
\begin{itemize}[leftmargin=*]
\item \emph{\model}$_{h}$: $P \in \{hour/30mins\}$
\item \emph{\model}$_{h,d}$: $P \in \{hour/30mins, day\}$
\item \emph{\model}$_{h,w}$: $P \in \{hour/30mins, week\}$
\item \emph{\model}$_{h,d,w}$: $P \in \{hour/30mins, day, week\}$
\end{itemize}

\begin{figure*}
	\centering
	\subfigure[][NYC-Taxi]{
		\centering
		\includegraphics[width=0.15\textwidth]{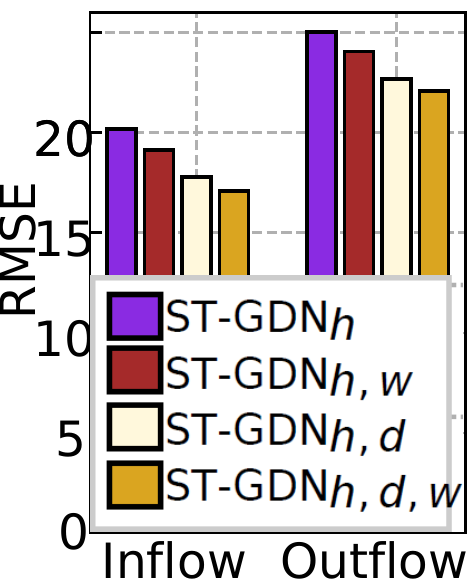}
		\label{fig:hdwtaxiNYC_RMSE}
	}
	\subfigure[][NYC-Taxi]{
		\centering
		\includegraphics[width=0.15\textwidth]{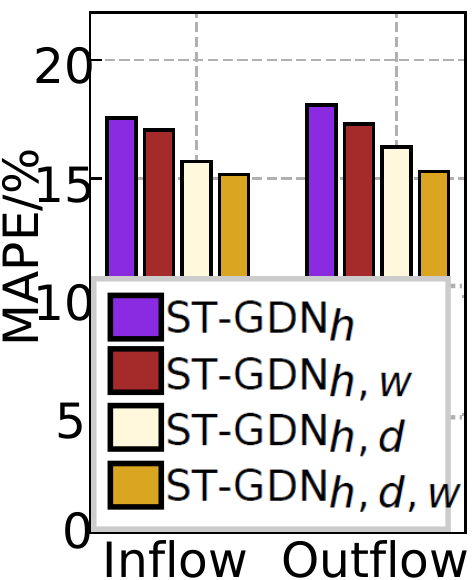}
		\label{fig:hdwtaxiNYC_MAPE}
	}
	\subfigure[][BJ-Taxi]{
		\centering
		\includegraphics[width=0.15\textwidth]{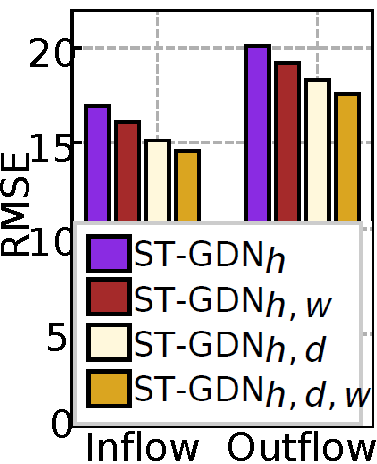}
		\label{fig:hdwtaxiBJ_RMSE}
	}
	\subfigure[][BJ-Taxi]{
		\centering
		\includegraphics[width=0.15\textwidth]{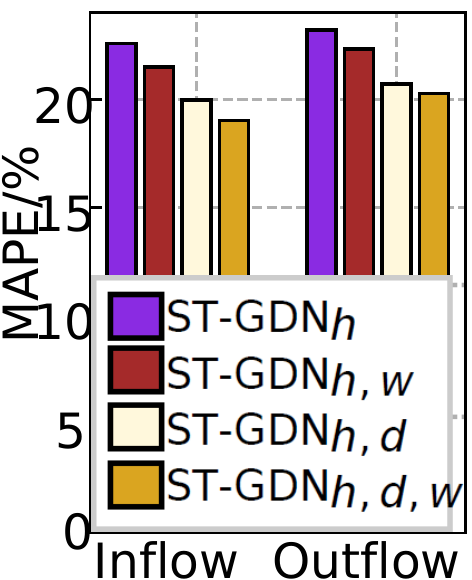}
		\label{fig:hdwtaxiBJ_MAPE}
	}
	\subfigure[][NYC-Bike]{
		\centering
		\includegraphics[width=0.15\textwidth]{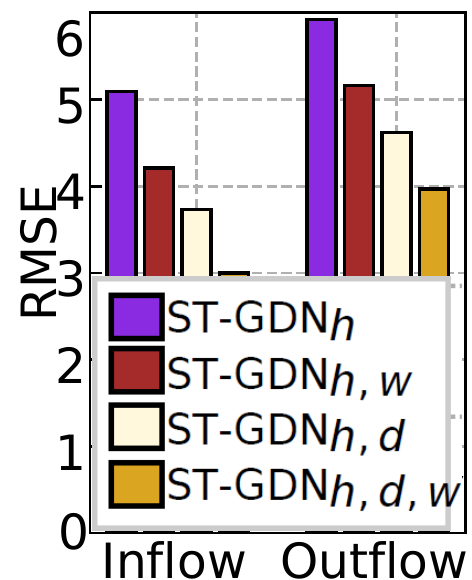}
		\label{fig:hdwbikeNYC_RMSE}
	}
	\subfigure[][NYC-Bike]{
		\centering
		\includegraphics[width=0.15\textwidth]{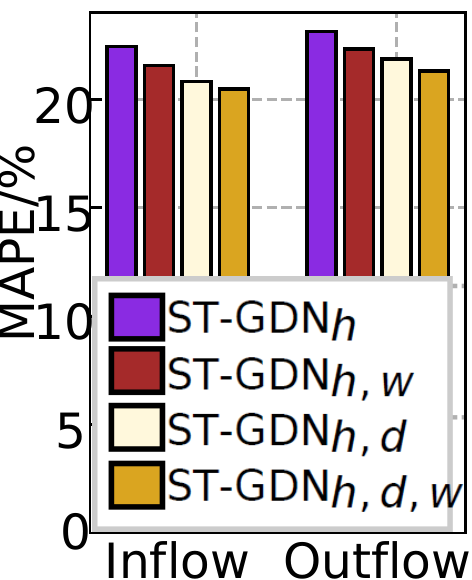}
		\label{fig:hdwbikeNYC_MAPE}
	}
	\vspace{-0.1in}
	\caption{Multi-resolution temporal effect studies.}
	\label{fig:hdw}
	\label{fig:resolution}
	\vspace{-0.05in}
\end{figure*}

We present the study results in Figure~\ref{fig:resolution}. As we can seen, the best prediction accuracy is achieved by \emph{\model}$_{h,d,w}$ which is configured with more resolutions. Leaning the temporal hierarchy with hourly and daily/weekly traffic patterns (\emph{\model}$_{h,d}$, \emph{\model}$_{h,w}$) provide better results as compared to the variant with singular-dimensional time granularity (\emph{\model}$_{h}$). Overall, decomposing the temporal effects into more multiple resolution-specific feature representations is helpful for more accurate modeling of traffic temporal regularity and resolution-aware region relations.

\begin{figure*}[t]
	\vspace{-0.1in}
	\centering
	\begin{adjustbox}{max width=1.0\linewidth}
		\begin{tikzpicture}
\begin{axis}[
xlabel={Filter Size},
ylabel={RMSE},
xmin=2,xmax=5,
ymin=15.5,ymax=24,
legend columns=2,
legend cell align=right,
grid=both,
every axis plot/.append style={ultra thick},
every tick label/.append style={scale=2.2},
label style={scale=3},
legend style={
	nodes={scale=2, transform shape},
	legend image post style={scale=2},
},
legend style={at={(0,0)},anchor=south west},
every axis plot post/.append style={
	every mark/.append style={scale=2}
}
]

\addplot[color={rgb:red,4;green,2;yellow,1}, mark=otimes, dashed]
table[x=para, y=End] {filtersizeRMSE.txt};
\addplot[color={rgb:blue,4;green,2;yellow,1}, mark=o, dashed, mark options={solid}]
table[x=para, y=Start] {filtersizeRMSE.txt};
\legend{\Large Inflow, \Large Outflow};

\end{axis}
\end{tikzpicture}

\begin{tikzpicture}
\begin{axis}[
xlabel={Hidden Dimensionality},
ylabel={RMSE},
xmin=32,xmax=256,
ymin=15.5,ymax=23.23,
legend columns=2,
legend cell align=right,
grid=both,
every axis plot/.append style={ultra thick},
every tick label/.append style={scale=2.2},
label style={scale=3},
legend style={
	nodes={scale=2, transform shape},
	legend image post style={scale=2},
},
legend style={at={(0,0)},anchor=south west},
every axis plot post/.append style={
	every mark/.append style={scale=2}
}
]

\addplot[color={rgb:red,4;green,2;yellow,1}, mark=otimes, dashed]
table[x=para, y=End] {filternumRMSE.txt};
\addplot[color={rgb:blue,4;green,2;yellow,1}, mark=o, dashed, mark options={solid}]
table[x=para, y=Start] {filternumRMSE.txt};
\legend{\Large Inflow, \Large Outflow};

\end{axis}
\end{tikzpicture}

\begin{tikzpicture}
\begin{axis}[
xlabel={Sequence Length $T_{h}$},
ylabel={RMSE},
xmin=1,xmax=6,
ymin=15.5,ymax=25,
legend columns=2,
legend cell align=right,
grid=both,
every axis plot/.append style={ultra thick},
every tick label/.append style={scale=2.2},
label style={scale=3},
legend style={
	nodes={scale=2, transform shape},
	legend image post style={scale=1.5},
},
legend style={at={(0,0)}, anchor=south west},
every axis plot post/.append style={
	every mark/.append style={scale=2}
}
]

\addplot[color={rgb:red,4;green,2;yellow,1}, mark=otimes, dashed]
table[x=para, y=End] {ncRMSE.txt};
\addplot[color={rgb:blue,4;green,2;yellow,1}, mark=o, dashed, mark options={solid}]
table[x=para, y=Start] {ncRMSE.txt};
\legend{\Large Inflow, \Large Outflow};

\end{axis}
\end{tikzpicture}

\begin{tikzpicture}
\begin{axis}[
xlabel={Sequence Length $T_{d}$ },
ylabel={RMSE},
xmin=1,xmax=5,
ymin=15.5,ymax=23.5,
legend columns=2,
legend cell align=right,
grid=both,
every axis plot/.append style={ultra thick},
every tick label/.append style={scale=2.2},
label style={scale=3},
legend style={
	nodes={scale=2, transform shape},
	legend image post style={scale=1.5},
},
legend style={at={(0,0)}, anchor=south west},
every axis plot post/.append style={
	every mark/.append style={scale=2}
}
]

\addplot[color={rgb:red,4;green,2;yellow,1}, mark=otimes, dashed]
table[x=para, y=End] {npRMSE.txt};
\addplot[color={rgb:blue,4;green,2;yellow,1}, mark=o, dashed, mark options={solid}]
table[x=para, y=Start] {npRMSE.txt};
\legend{\Large Inflow, \Large Outflow};

\end{axis}
\end{tikzpicture}

\begin{tikzpicture}
\begin{axis}[
xlabel={Sequence Length $T_{w}$},
ylabel={RMSE},
xmin=1,xmax=3,
ymin=15.5,ymax=23,
legend columns=2,
legend cell align=right,
grid=both,
every axis plot/.append style={ultra thick},
every tick label/.append style={scale=2.2},
label style={scale=3},
legend style={
	nodes={scale=2, transform shape},
	legend image post style={scale=1.5},
},
legend style={at={(0,0)}, anchor=south west},
every axis plot post/.append style={
	every mark/.append style={scale=2}
}
]

\addplot[color={rgb:red,4;green,2;yellow,1}, mark=otimes, dashed]
table[x=para, y=End] {ntrdRMSE.txt};
\addplot[color={rgb:blue,4;green,2;yellow,1}, mark=o, dashed, mark options={solid}]
table[x=para, y=Start] {ntrdRMSE.txt};
\legend{\Large Inflow, \Large Outflow};

\end{axis}
\end{tikzpicture}

\begin{tikzpicture}
\begin{axis}[
xlabel={Number of GAT Layers},
ylabel={RMSE},
xmin=1,xmax=5,
ymin=15.5,ymax=23,
legend columns=2,
legend cell align=right,
grid=both,
every axis plot/.append style={ultra thick},
every tick label/.append style={scale=2.2},
label style={scale=3},
legend style={
	nodes={scale=2, transform shape},
	legend image post style={scale=1.5},
},
legend style={at={(0,0)},anchor=south west},
every axis plot post/.append style={
	every mark/.append style={scale=2}
}
]

\addplot[color={rgb:red,4;green,2;yellow,1}, mark=otimes, dashed]
table[x=para, y=End] {aRMSE.txt};
\addplot[color={rgb:blue,4;green,2;yellow,1}, mark=o, dashed, mark options={solid}]
table[x=para, y=Start] {aRMSE.txt};
\legend{\Large Inflow, \Large Outflow};

\end{axis}
\end{tikzpicture}
	\end{adjustbox}
	\vspace{-0.1in}
	\caption{Hyper-parameter study on NYC-Taxi data in terms of RMSE.}
    \vspace{-0.1in}
	\label{fig:hyperparameter}
\end{figure*}
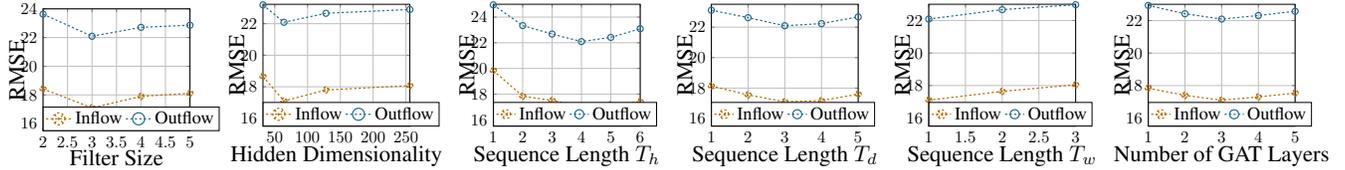

% \subsection{Qualitative Analyses: Model Interpretation (RQ4)}
% Besides the performance gain, we show the explainability of \emph{\model} by visualizing the attentive relevance scores learned from our graph-based attention network for capturing global traffic pattern dependencies across regions. As illustrated in Figure~\ref{fig:case_study}, it is worthwhile pointing out that the target region (``Dongzhimen Bridge'') in Beijing are highly relevant to other regions which are either spatially adjacent areas, or share the similar functions (\eg, shopping zone and transportation hub). This observation again confirms the effectiveness of \emph{\model}
% in capturing both local and global inter-dependencies across different regions. Furthermore, a side benefit of \emph{\model} is that it could provide explicit and reasonable explanations for traffic prediction results.

% \begin{figure}[h!]
%     \centering
%     
%     \includegraphics[width=0.46\textwidth]{images2020/weightmap_2.pdf}
%     
%     \caption{Case study from BJ-Taxi data. The target region (Dongzhimen Bridge) has three key urban functions (\ie, CBD, Transportation Hub, Subway Station, Shopping Center). We highlight eight highly relevant regions which share similar geographical functions with the target region.}
%     \label{fig:case_study}
%     
% \end{figure}

\subsection{Parameter Sensitivity (RQ4)}

\noindent \textbf{Depth of Graph Attention Network} $L$. We can notice that increasing the depth of our graph attention module by stacking multiple embedding propagation layers could boost the performance. The results also indicate that exploring third-order relations among region entities is sufficient to capture the global traffic dependencies.

%We change the depth our graph attention module to examine the efficacy of stacking multiple embedding progation layers

\noindent \textbf{Length of Encoded Input Sequence} $T$. The performance is initially improved with the increase of $T_h$ and $T_d$, since longer traffic series can provide more useful temporal information. However, the further increasing of sequence length may introduce noise which mislead the traffic modeling.\\\vspace{-0.12in}

\noindent \textbf{Kernel Size} $K$. We vary the kernel size to investigate the convolution operations in our graph diffusion process. We can observe that $K=3$ achieves the best performance.\\\vspace{-0.12in}

\noindent \textbf{Channel Dimensionality}. The results suggest that larger channel dimension size does not always bring the stronger representation ability, due to the overfitting issue.

\subsection{Model Efficiency Study (RQ5)}
We finally investigate the model efficiency (measured by running time) of our \textit{\model}. All experiments are conducted with the default parameter configurations on a single NVIDIA GeForce GTX 1080 Ti GPU. We observe that in several best performed baselines, \emph{ST-GCN} has good prediction accuracy and running speed. Our \emph{\model} outperforms most of compared approaches and could achieve competitive efficiency as compared to \emph{ST-GCN}, \ie, the attention-based graph embedding propagation layer has higher computational cost than the adjacent matrix-based graph convolution. Considering the prediction accuracy comparison between \emph{\model} and \emph{ST-GCN}, the additional computational cost could bring positive effect via learning global region inter-dependencies in an explicit manner.

\begin{table}%[htbp]
    \centering
    \normalsize
    %\scriptsize
	\label{tab: running time}
	\begin{tabular}{|c|ccc|}
		\hline
		\multirow{2}*{Methods}&
		\multicolumn{3}{c|}{Training}\\
		\cline{2-4}
		&BJ-Taxi&NYC-Taxi&NYC-Bike\\
		\hline
		ST-MetaNet&16121.01&1298.55&1020.14\\
		DCRNN&7996.24&981.36&705.64\\
		ST-GCN&4088.90&744.65&500.40\\
		ST-MGCN&8263.27&1023.29&789.71\\
		GMAN&7368.31&854.66&547.12\\
		\hline
		\emph{\model}&7625.19&891.63&569.26\\
		\hline
		\multirow{2}*{Methods}& \multicolumn{3}{c|}{Inference}\\
		\cline{2-4}
		&BJ-Taxi&NYC-Taxi&NYC-Bike\\
		\hline
		ST-MetaNet&0.42&0.32&0.29\\
		DCRNN&0.26&0.24&0.21\\
		ST-GCN&0.25&0.22&0.19\\
		ST-MGCN&0.31&0.27&0.25\\
		GMAN&0.25&0.23&0.19\\
		\hline
		\emph{\model}&0.26&0.23&0.20\\
		\hline
	\end{tabular}
	\caption{Model Efficiency Study.}
	\
	\label{tab:time}
\end{table}

We finally investigate the efficiency (measured by running time) of our \textit{\model}. Table~\ref{tab:time} presents the computational cost of training (with 300 epochs) and inference phase for \textit{\model} and five best performed baselines on three different datasets. All experiments are conducted with the default parameter configurations on a single NVIDIA GeForce GTX 1080 Ti GPU. We can observe that \emph{\model} outperforms most of compared approaches and could achieve competitive efficiency as compared to \emph{ST-GCN}, \ie, the attention-based graph embedding propagation layer has higher computational cost than the adjacent matrix-based graph convolution. Considering the prediction accuracy comparison between \emph{\model} and \emph{ST-GCN}, the additional computational cost could bring positive effect via learning global region inter-dependencies in an explicit manner.

%\fi
\section{Related Work}
\label{sec:relate}

\noindent \textbf{Traffic Prediction with Deep Learning}. Recently, many efforts have been devoted to developing traffic prediction techniques based on various neural network architectures. One straightforward solution is to apply the recurrent neural networks (\eg, LSTM) to encode the temporal features of traffic series~\cite{yu2017deep}. The subsequent extensions propose to integrate the recurrent neural layers with the convolutional network~\cite{zhang2017deep,yao2018deep} or attention mechanism~\cite{yao2019revisiting}, so as to joint model the spatial-temporal signals. In addition, some hybrid methods have been proposed for traffic prediction with the exploration of heterogeneous data fusion~\cite{liang2019urbanfm} and meta-learning-based knowledge transfer~\cite{pan2019urban}. Different from these work, \model\ endows the spatial-temporal pattern representation process with the preservation of hierarchical temporal dynamics and global-enhanced region-wise dependencies. While there exist research work that considers the global dependency among regions~\cite{zhang2020spatial}, it is limited in its separately modeling of traffic dependency and nearby region relations based on convolution neural network. In this work, \model\ incorporates the global context enhanced region-wise explicit relevance into a graph diffusion paradigm to capture comprehensive high-order region dependencies in a joint learning manner. \\

\noindent \textbf{Graph-based Spatial-Temporal Prediction}. It is worth mentioning that several recent efforts have investigated Graph Neural Networks (GNNs) for spatial-temporal data forecasting~\cite{guo2019attention,song2020spatial}. For example, ST-GCN~\cite{yubingspatio} and ST-MGCN~\cite{geng2019spatiotemporal} proposes to leverage graph convolution network to model correlations between regions. Furthermore, attention mechanism has been introduced for information aggregation from adjacent roads~\cite{zheng2020gman}. Motivated by these work, we develop a hierarchical graph neural architectures to promote the cooperation between the multi-resolution temporal context with the cross-region inter-correlations, which have not been well explored in existing solutions. 
\section{Conclusion}
\label{sec:conclusion}

This work investigates the traffic prediction problem by proposing a new architecture (\model) based graph neural networks. Specifically, it first designs a resolution-aware self-attention network to encode the multi-level temporal signals. Then, the local spatial contextual information and global traffic dependencies across different regions, are subsequently integrated to enhance the spatial-temporal pattern representations. Comprehensive experiments demonstrate that the proposed \model\  significantly outperforms many baselines over several datasets consistently. Our future work lies in the deployment of our developed prototype in a cloud-based working system for real-time traffic prediction.

\section*{Acknowledgments}
The authors would like to thank the anonymous referees for their valuable comments and helpful suggestions. This work is supported by National Nature Science Foundation of China (62072188, 61672241), Natural Science Foundation of Guangdong Province (2016A030308013), Science and Technology Program of Guangdong Province (2019A050510010). This work is also partially supported by National Key R\&D Program of China (2019YFB2101801) and the Beijing Nova Program (Z201100006820053).\vspace{-0.1in}

%\clearpage

% \bibliographystyle{aaai}
\bibliography{refs}

\end{document}